\documentclass[11pt]{article}

\usepackage{preprint}

\newif\ifdraftnotes
\draftnotestrue

\definecolor{RoyalBlue}{RGB}{0,100,170}
\definecolor{EasternBlue}{RGB}{37,150,190}
\definecolor{Cerulean}{RGB}{80,150,220}
\definecolor{Emerald}{RGB}{62,156,94}
\definecolor{Orange}{RGB}{250,150,50}
\definecolor{Rouge}{RGB}{250,95,95}
\definecolor{peach}{rgb}{1,0.56,0.56}
\definecolor{coral}{RGB}{240,128,128}
\definecolor{sand}{RGB}{250,150,120}
\definecolor{grass}{RGB}{120,190,50}
\definecolor{sky}{RGB}{50,150,250}
\definecolor{midgray}{RGB}{150,150,150}

\definecolor{SpellPurple}{HTML}{6A3D9A}   
\definecolor{MageGreen}{HTML}{1B7837}     
\definecolor{PCAPink}{HTML}{E7298A}       
\definecolor{RandomBlue}{HTML}{2B6CB0}    
\definecolor{OracleInk}{HTML}{111111}     
\definecolor{SphSVDTeal}{HTML}{004449}    
\definecolor{FirstOrange}{HTML}{D95F02}   

\definecolor{ColorDef}{RGB}{80,180,150}
\definecolor{RevisionRed}{RGB}{240,35,35}
\definecolor{RevisionBlue}{RGB}{80,180,250}
\definecolor{TODOorange}{RGB}{255,150,50}

\hypersetup{
  colorlinks = true,
  linkcolor  = RoyalBlue,
  citecolor  = RoyalBlue,
  urlcolor   = RoyalBlue,
  breaklinks = true,
}

\lstdefinestyle{plain}{
  basicstyle       = \ttfamily\small,
  keywordstyle     = \color{RoyalBlue}\bfseries,
  commentstyle     = \color{midgray}\itshape,
  stringstyle      = \color{Emerald},
  numberstyle      = \tiny\color{midgray},
  numbers          = left,
  breaklines       = true,
  showstringspaces = false,
  frame            = none,
  columns          = fullflexible,
}
\NewDocumentCommand{\DeclareAuthorNote}{ m m m }{%
  \definecolor{#1notecolour}{HTML}{#3}%
  \expandafter\newcommand\csname #1\endcsname[1]{%
    \ifdraftnotes
      \textcolor{#1notecolour}{\textsf{\small[#2: ##1]}}%
    \fi
  }%
}

\DeclareAuthorNote{ericnote}{Eric}{1F77B4}
\DeclareAuthorNote{andrewnote}{Andrew}{2CA02C}
\DeclareAuthorNote{arinote}{Ari}{B027D6}

\makeatletter
\@ifpackageloaded{ulem}
  {}
  {}
\makeatother

\theoremstyle{plain}

\theoremstyle{definition}

\theoremstyle{remark}

\theoremstyle{plain}
\newtheorem*{theorem*}{Theorem}
\newtheorem*{lemma*}{Lemma}
\newtheorem*{proposition*}{Proposition}
\newtheorem*{corollary*}{Corollary}
\newtheorem*{claim*}{Claim}
\newtheorem*{question*}{Question}
\theoremstyle{definition}
\newtheorem*{definition*}{Definition}
\theoremstyle{remark}
\newtheorem*{remark*}{Remark}

\theoremstyle{plain}
\newcommand{\theoremname}{Theorem}
\newtheorem*{namedtheorem}{\theoremname}

\theoremstyle{plain}

\crefname{assumption}{Assumption}{Assumptions}
\Crefname{assumption}{Assumption}{Assumptions}
\crefname{condition}{Condition}{Conditions}
\Crefname{condition}{Condition}{Conditions}
\crefname{observation}{Observation}{Observations}
\Crefname{observation}{Observation}{Observations}
\crefname{claim}{Claim}{Claims}
\Crefname{claim}{Claim}{Claims}
\crefname{example}{Example}{Examples}
\Crefname{example}{Example}{Examples}

\def\Figref#1{Figure~\ref{#1}}

\def\Secref#1{Section~\ref{#1}}

\def\eqref#1{equation~\ref{#1}}
\def\Eqref#1{Equation~\ref{#1}}

\def\Algref#1{Algorithm~\ref{#1}}

\allowdisplaybreaks

\newlength{\keyrulewd}
\newlength{\keyraise}
\newcommand{\keybox}[1]{\makebox[9.5pt][c]{#1}}
\newcommand{\keyglyph}[2]{\textcolor{#1}{$\scriptstyle #2$}}
\newcommand{\keyseg}[2]{\textcolor{#1}{\rule[\keyraise]{#2}{\keyrulewd}}}
\newcommand{\keygap}{\hspace{1.3pt}}
\newcommand{\keydash}[1]{\keyseg{#1}{3.2pt}\keygap\keyseg{#1}{3.2pt}}
\newcommand{\keydot}[1]{\keyseg{#1}{0.9pt}\keygap\keyseg{#1}{0.9pt}%
                        \keygap\keyseg{#1}{0.9pt}\keygap\keyseg{#1}{0.9pt}}
\newcommand{\keydashdot}[1]{\keyseg{#1}{3.4pt}\keygap\keyseg{#1}{0.9pt}%
                            \keygap\keyseg{#1}{2.4pt}}

\newcommand{\kMage}{\keybox{\keyglyph{MageGreen}{\blacktriangle}}}
\newcommand{\kSpell}{\keybox{\keyglyph{SpellPurple}{\bullet}}}
\newcommand{\kRandom}{\keybox{\keyglyph{RandomBlue}{\blacktriangledown}}}
\newcommand{\kPCA}{\keybox{\keyglyph{PCAPink}{\blacklozenge}}}
\newcommand{\kSVD}{\keybox{\keydot{OracleInk}}}
\newcommand{\kSphSVD}{\keybox{\keydashdot{SphSVDTeal}}}
\newcommand{\kFirst}{\keybox{\keyglyph{FirstOrange}{\bigstar}}}
\newcommand{\kMagic}{\keybox{\keydash{OracleInk}}}

\newcolumntype{R}{>{$}r<{$}}

\newcommand{\se}[1]{\,\mbox{\scriptsize$\pm#1$}}

\newcommand{\best}[1]{\text{\textbf{#1}}}

\DeclarePairedDelimiterX{\inner}[2]{\langle}{\rangle}{#1,#2}

\DeclareMathOperator{\diag}{diag}

\makeatletter
\@ifpackageloaded{bbm}{}{}
\makeatother
\def\1{\bm{1}}

\DeclareMathAlphabet{\mathsfit}{\encodingdefault}{\sfdefault}{m}{sl}
\SetMathAlphabet{\mathsfit}{bold}{\encodingdefault}{\sfdefault}{bx}{n}

  \def\gL{{\mathcal{L}}}

\NewDocumentCommand{\loss}{o}{\ensuremath{\ell\IfValueT{#1}{^{(#1)}}}}
\NewDocumentCommand{\Loss}{o}{\ensuremath{\gL\IfValueT{#1}{^{(#1)}}}}
\NewDocumentCommand{\lr}{o}{\ensuremath{\eta\IfValueT{#1}{^{(#1)}}}}  

\NewDocumentCommand{\param}{o}{\ensuremath{\theta\IfValueT{#1}{^{(#1)}}}}
\NewDocumentCommand{\Param}{o}{\ensuremath{\theta\IfValueT{#1}{^{(#1)}}}}
\NewDocumentCommand{\tparam}{o}{\ensuremath{\tilde{\theta}\IfValueT{#1}{^{(#1)}}}}
\NewDocumentCommand{\grad}{o}{\ensuremath{g\IfValueT{#1}{^{(#1)}}}}
\NewDocumentCommand{\gradrot}{o}{\ensuremath{\tilde{g}\IfValueT{#1}{^{(#1)}}}}
\NewDocumentCommand{\Grad}{o}{\ensuremath{G\IfValueT{#1}{^{(#1)}}}}
\NewDocumentCommand{\DeltaParam}{o}{\ensuremath{\Delta\IfValueT{#1}{^{(#1)}}}}
\NewDocumentCommand{\update}{o}{\ensuremath{\delta\IfValueT{#1}{^{(#1)}}}}

\NewDocumentCommand{\paramCov}{o}{\ensuremath{M\IfValueT{#1}{^{(#1)}}}}
\NewDocumentCommand{\tparamCov}{o}{\ensuremath{\widetilde{M}\IfValueT{#1}{^{(#1)}}}}
\NewDocumentCommand{\precond}{o}{\ensuremath{P\IfValueT{#1}{^{(#1)}}}}
\NewDocumentCommand{\precondD}{o}{\ensuremath{D\IfValueT{#1}{^{(#1)}}}}

\algrenewcommand\algorithmicrequire{\textbf{Input:}}
\algrenewcommand\algorithmicensure{\textbf{Output:}}

\makeatletter
\newcommand{\appropto}{\mathrel{\mathpalette\appropto@\relax}}
\newcommand{\appropto@}[2]{%
  \begingroup
  \setbox\z@=\hbox{$\m@th#1\propto$}
  \hbox{\ooalign{%
    \hfil\kern-0.06\wd\z@\raise0.46\ht\z@\hbox{$\m@th#1\propto$}\hfil\cr
    \hfil\lower0.52\ht\z@\hbox{$\m@th#1\sim$}\hfil\cr
  }}%
  \endgroup}
\makeatother

\title{Data Attribution via Sketched Metadifferentiation}
\author{Yuxi Chen$^{1}$, Hamza Golubovic$^{2}$, Han Tong$^{2}$, Arian Maleki$^{2, 3}$, and Andrew Ilyas$^{4, 5}$\\\\[2pt]
Carnegie Mellon University\\
$^{1}$Department of Statistics \& Data Science\\
$^{4}$Software and Societal Systems Department\\
$^{5}$Department of Electrical and Computer Engineering\\
\texttt{\{ericc3, andrewi\}@andrew.cmu.edu}\\\\
Columbia University\\
$^{2}$Department of Statistics\\
$^{3}$Electrical Engineering Department\\
\texttt{\{hg2723, ht2672, mm4338\}@columbia.edu}
}
\date{\today}

\begin{document}
\maketitle

\begin{abstract}
Data attribution seeks to quantify how individual training examples shape a model's predictions and underpins problems including data valuation, machine unlearning, and model interpretability. Despite having a long line of work, computationally scalable methods often struggle to predict the effect of removing training data in neural networks due to their non-convex nature. To overcome this challenge, metagradient-based methods such as MAGIC \citep{ilyas_magic_2025} differentiate each prediction through the entire training run and compute its exact influence with respect to the training data, but require a separate run for every prediction. To reduce this cost, we cast budgeted attribution as estimating a large influence matrix from a small number of measurements. We show that the measurements most appropriate for recovering this matrix differ from those best suited for attribution itself. We then present two algorithms, \textsc{Mage} and \textsc{Spell}, suited for reconstruction and attribution respectively, that run on existing metagradient machinery at no extra cost. Empirical studies demonstrate strong performance over existing baselines across training scales and measurement budgets.
\end{abstract}

\section{Introduction}

With the success of modern deep learning, understanding how training data shapes model behavior has become increasingly important. Previous studies have traced many undesirable model behaviors back to the training data, showing that neural networks can memorize private records \citep{carlini_extracting_2021}, inherit targeted vulnerabilities from a few poisoned examples \citep{wan_poisoning_2023}, and learn shortcuts by relying on signals that are predictive yet brittle \citep{ilyas_adversarial_2019}. At the same time, data selection allows smaller datasets to match or surpass the performance of larger counterparts for vision and language model training \citep{sorscher_beyond_2022, li_datacomp-lm_2024}. Similar effects likewise appear during post-training, where a curated fraction of the data can elicit strong instruction following and reasoning \citep{zhou_lima_2023, xia_less_2024, muennighoff_s1_2025}. Together, these findings testify to the central role of training data in shaping model behavior.

One way to understand the impact of training data is to ask how a given metric of interest (e.g., held-out test loss) would have changed had the same model been trained on a different dataset. To answer this question naively would require one to modify the training dataset, retrain the new model, and observe the resulting outputs. Instead, predictive data attribution seeks to predict this counterfactual quantity without retraining \citep{ilyas_datamodels_2022} by estimating the contribution of every training example to the prediction of interest. Previous works have studied this problem using tools like approximate influence functions \citep{koh_understanding_2017}, Shapley values \citep{ghorbani_data_2019}, trajectory tracing \citep{pruthi_estimating_2020}, and scalable approximations of influence functions \citep{park_trak_2023}. Despite their computational appeal, these methods have often produced estimates that align poorly with the true outcome of retraining in neural networks \citep{bae_if_2022, bae_training_2024}.

Overcoming these hurdles, MAGIC \citep{ilyas_magic_2025} differentiates the loss from a test prediction backward through the entire optimization trajectory, computing its gradient with respect to the training data weights. This gradient---also known as the \emph{exact influence function}---near-perfectly predicts the effects of modifying the training dataset on the given test prediction's loss. However, computing the exact influence function with MAGIC would require a separate reverse differentiation through the full training run for every prediction of interest. This makes MAGIC expensive to scale to settings where we wish to study many held-out examples individually.

In this work, we ask whether the cost of data attribution can be reduced when estimating the influence scores across many predictions. Our contributions are as follows:
\begin{enumerate}
    \item We cast the problem of budgeted data attribution across many predictions as that of estimating a Jacobian matrix of influence scores from a limited number of left and right vector products.
    \item We formalize two distinct objectives, reconstruction and attribution, and show that the two are optimized using different measurements, since reconstruction weights each prediction by its squared row norm whereas attribution weights them equally.
    \item We introduce \textsc{Mage} (\textbf{M}etagradient \textbf{A}pproximation by \textbf{G}ram \textbf{E}igendirections) for reconstruction and \textsc{Spell} (\textbf{SP}herical \textbf{E}stimation for the \textbf{L}inear datamodeling score at \textbf{L}ow budgets) for attribution, both of which run on existing metagradient machinery.
    \item Across a ResNet-9 on CIFAR-10 and a 92M-parameter language model on TinyStories, \textsc{Mage} and \textsc{Spell} each outperform existing computable alternatives on their intended objective under various measurement budgets.
\end{enumerate}

\section{Predictive Data Attribution and Influence Matrix Estimation} 

We begin by formalizing both the predictive data attribution framework and the underlying computational primitives. The central idea is to parameterize the training dataset through continuous importance weights, after which our goal becomes computing derivatives of the trained model behavior with respect to these weights. Finally, we state the problem of estimating the influence matrix under a limited budget and define two objectives, reconstruction and attribution, which emphasize distinct aspects of any estimate.

\subsection{Setup and Notation}

We follow the datamodeling framework of \citet{ilyas_datamodels_2022}. Let $\mathcal{D}=\{x_i\}_{i=1}^n$ be a dataset of $n$ training points. We parameterize the training set by an importance-weight vector $w\in\mathbb{R}^n$, where $w_i$ scales the contribution of example $i$ to the training loss. With this representation, a standard training pipeline corresponds to $w=\mathbf{1}_n$, where $\mathbf{1}_n\in\mathbb{R}^n$ is the all-ones vector, and setting $w_i=0$ amounts to removing example $i$ from the training set. 

Next, let $\Theta\subseteq\mathbb{R}^p$ denote the parameter space and let $\mathcal{A}$ be a learning algorithm mapping data weights to trained model parameters,
\[
\mathcal{A}:\mathbb{R}^n\rightarrow\Theta,\qquad \theta(w)=\mathcal{A}(w).
\]
The learning algorithm $\mathcal{A}$ encompasses every aspect of training beyond the data weights, including the model architecture, optimizer, and learning rate schedule.

Following the single-model setting of \citet{ilyas_magic_2025}, we also fix every source of training randomness, whereby the same weighting $w$ always generates the same parameters $\theta(w)$. We further assume that $\mathcal{A}$ is iterative and smooth, meaning that $\theta(w)$ is produced by a fixed sequence of differentiable training updates, and small perturbations to $w$ produce controlled changes in the resulting model output and its derivative. 

Finally, we are given $k$ test queries that we wish to attribute to the training data. Throughout, we write $[k]=\{1,\ldots,k\}$. For each test query $q\in[k]$, we let \(\phi_q:\Theta\rightarrow\mathbb{R}\) be a differentiable scalar measurement of the trained model (e.g., the loss of the model with parameters $\theta$ on test query $q$). We define the model output function $f_q$ 
\begin{equation}
f_q(w)=\phi_q(\theta(w))=\phi_q(\mathcal{A}(w))
\label{eq:composition}
\end{equation}
as the composition function mapping data weights directly to the measurement for query $q$, so that $f_q(w)$ is the value that would be observed on query $q$ if the model were trained using data weights $w$. We collect these functions into the vector-valued model output function $F$,
\[
F(w)=\left(f_1(w),\ldots,f_k(w)\right)\in\mathbb{R}^k.
\]
At a high level, our goal is to return an estimate $\widehat{F}=(\widehat{f}_1,\ldots,\widehat{f}_k)$ of this function, which predicts the model behavior under any weighting $w$ without retraining.

\subsection{Metagradients and MAGIC}

The composition in \Eqref{eq:composition} makes the model output function $f_q(w)$ differentiable with respect to the training weights, and its derivative
\[
\nabla_w f_q(\mathbf{1}_n)
=\nabla_w\,\phi_q(\mathcal{A}(w))\big|_{w=\mathbf{1}_n}\in\mathbb{R}^n
\]
is called the \emph{metagradient} of query $q$, a gradient taken through the entire training procedure. Its $i$th entry measures how the observed value on query $q$ would respond to an infinitesimal change in the weight of training example $i$. Since computing this derivative naively would be computationally prohibitive, \citet{engstrom_optimizing_2025} introduce REPLAY, which computes metagradients for large-scale training runs at a cost comparable to that of training itself. Hereinafter, we use the word ``replay'' synonymously with the REPLAY algorithm.

To leverage metagradients for data attribution, MAGIC \citep{ilyas_magic_2025} linearizes the model output function $f_q$ around \(w = \mathbf{1}_n\) with a first-order Taylor expansion,
\[
f_q(w)\approx\widehat{f}_q(w)=f_q(\mathbf{1}_n)+\langle\nabla_w f_q(\mathbf{1}_n),\,w-\mathbf{1}_n\rangle,
\]
which defines a linear predictor for studying the effect of removing any subset of training examples. Let $S\subseteq[n]$ be a subset of examples to remove and let $\mathbf{1}_S\in\{0,1\}^n$ be its indicator vector, so that training without the examples in $S$ corresponds to the weighting $\mathbf{1}_n-\mathbf{1}_S$. Writing $\Delta_q(S)$ for the resulting change in query $q$, MAGIC predicts
\[
\Delta_q(S)=f_q(\mathbf{1}_n-\mathbf{1}_S)-f_q(\mathbf{1}_n)
\approx-\left\langle\nabla_w f_q(\mathbf{1}_n),\,\mathbf{1}_S\right\rangle
=-\sum_{i\in S}\left[\nabla_w f_q(\mathbf{1}_n)\right]_i.
\]
Since the metagradient is evaluated at $\mathbf{1}_n$ and does not depend on $\mathbf{1}_S$, a single metagradient allows us to generate counterfactual predictions for all subsets of \(\mathcal{D}\) simultaneously, and these predictions have been shown to near-perfectly estimate the effects of removing training data \citep{ilyas_magic_2025}.

Finally, we collect the metagradients across all $k$ queries and define the \emph{influence matrix}
\[
Y=\left.\frac{\partial F(w)}{\partial w}\right|_{w=\mathbf{1}_n}\in\mathbb{R}^{k\times n},
\]
which records the influence of every training example on every test query. We write $Y_q$ for its $q$th row---the metagradient of query $q$---so that recovering $Y$ would allow us to generate counterfactual predictions for all $k$ queries simultaneously.

\subsection{Influence Matrix Estimation}
\label{subsec:objectives}

MAGIC recovers row $Y_q$ by differentiating $f_q$ backward through the entire training trajectory. Thus, recovering all $k$ rows would require $k$ replays of training. In this paper, we ask:
\begin{center}
\textit{Given a budget of only $B < k$ replays, how well can we approximate the influence matrix?}
\end{center}
We consider two ways of evaluating the quality of any estimate $\widehat{Y}$ of the full influence matrix under this setting, which emphasize distinct yet equally important aspects of the problem.

\textbf{Reconstruction.} The reconstruction error is given by 
\[
\left\|Y-\widehat{Y}\right\|_F^2=\sum_{q=1}^k\left\|Y_q-\widehat{Y}_q\right\|^2,
\]
which is the standard measure of estimation quality for matrix recovery. This is an appropriate objective in settings where we wish to identify the training data driving model behavior, as is the case with data valuation \citep{ghorbani_data_2019, jia_towards_2019}.

\textbf{Attribution.} Instead of optimizing directly for the proximity between $\widehat{Y}$ and $Y$ in \(\ell_2\)-norm, we can also measure how well the induced predictions track the behavior of models retrained on the modified data. One objective that captures this notion of counterfactual accuracy is the linear datamodeling score (LDS) \citep{ilyas_datamodels_2022}. To define the LDS for a given query \(q\) at removal fraction \(p\), we draw a subset $S\subseteq[n]$ of size $np$ uniformly at random and write
\[
\widehat{\Delta}_q(S)=-\langle\widehat{Y}_q,\mathbf{1}_{S}\rangle,
\qquad
\Delta_q(S)=f_q(\mathbf{1}_n-\mathbf{1}_{S})-f_q(\mathbf{1}_n)
\]
for the predicted change and the true change observed on the model retrained using $[n]\setminus S$. The LDS for query $q$ is then given by
\[
\mathrm{LDS}_q=\rho_S\left(\widehat{\Delta}_q(S),\Delta_q(S)\right),
\]
where $\rho_S$ is the (Spearman) correlation taken over the randomness of $S$, and we average the LDS over all the queries. This is an appropriate objective in settings where we wish to predict counterfactual model behaviors under interventions on the training data, as in the case of machine unlearning \citep{guo_certified_2020, bourtoule_machine_2021} and model interpretability \citep{koh_understanding_2017}.

\section{\textsc{Mage} and \textsc{Spell}}

In this section, we present \textsc{Mage} and \textsc{Spell}, our procedures for estimating the influence matrix $Y$ from a budget of $B < k$ calls to the metagradient machinery. The two methods share a conceptually straightforward skeleton of measuring $B$ well-chosen linear combinations of the rows of $Y$, and estimating every row by its projection onto the span of the measurements. They differ primarily in the rule selecting the next combination. We present both procedures in \Algref{alg:mage-spell}.

\subsection{One Replay Can Measure Combinations of Queries}
\label{sec:method-oracle}

Recall that MAGIC calculates the attribution scores of a single query using reverse-mode differentiation through the training run. However, since differentiation is linear, MAGIC's computation is not constrained to a single query. To make this precise, let
\[
J=\left.\frac{\partial\theta(w)}{\partial w}\right|_{w=\mathbf{1}_n}\in\mathbb{R}^{p\times n}
\]
denote the Jacobian of the trained model parameters with respect to the $n$ training weights, whose $i$th column represents how perturbing the weight of training example $i$ changes the resulting model parameters. Next, for each query $q$, let
\[
v_q=\nabla_\theta\,\phi_q(\theta(\mathbf{1}_n))\in\mathbb{R}^p
\]
denote the gradient of its measurement with respect to the final parameters. By the chain rule, the influence matrix factors as
\[
Y=V^\top J\in\mathbb{R}^{k\times n}, \qquad V=[v_1,\ldots,v_k]\in\mathbb{R}^{p\times k}.
\]
Ordinarily, we would recover row $Y_q$ by running REPLAY with $v_q$ as the input to reverse-mode differentiation. More generally, for any coefficient vector $z$, we may instead input the combined query gradient $Vz$, where the same computation returns
\[
(Vz)^\top J=z^\top V^\top J=z^\top Y\in\mathbb{R}^n .
\]
Therefore, a single replay can measure any chosen linear combination of the rows of $Y$, and a budget of $B$ replays allows for $B$ such chosen combinations.

\subsection{A Projection Estimator via Forward-Mode Differentiation}
\label{subsec:method-estimator}

Having laid out the above, we can estimate every row of $Y$ by its projection onto the span of the acquired measurements. Writing
\[
\mathcal{U}=\operatorname{span}\left(z_1^\top Y,\ldots,z_B^\top Y\right)\subseteq\mathbb{R}^n
\]
for the subspace derived from \(B\) combinations $z_1,\ldots,z_B$, one natural candidate is  $\widehat{Y}=Y\Pi_{\mathcal{U}}$, where $\Pi_{\mathcal{U}}$ denotes the orthogonal projection onto $\mathcal{U}$.

However, we note that replays themselves can not determine the projection onto \(\mathcal{U}\). This is because projecting row $Y_q$ onto $\mathcal{U}$ also requires its inner product with every measurement, 
\[
\left\langle Y_q,\,z^\top Y\right\rangle=v_q^\top J\left(z^\top Y\right)^\top,
\]
which involves the unknown row itself. Despite this, we note that the factor $J(z^\top Y)^\top\in\mathbb{R}^p$ is common to every query, and can be calculated using standard forward-mode differentiation, as observed by \citet{gunn_how_2026}. We then pair each replay with an extra forward-mode pass to obtain
these inner products. Having this information allows us to update the estimate along the part of the new measurement orthogonal to all previous replays \citep{golub_matrix_2013}, a step we call \textsc{Extend-Projection} in \Algref{alg:mage-spell}.

\subsection{Reducing the Two Objectives}
\label{subsec:reduction}

To design our measurement strategy, we first note that the nature of orthogonal projections paves the way for powerful reductions of both objectives, which we derive below and make use of throughout our procedures.

\textbf{Reconstruction.}
For any query $q$, we can expand the reconstruction error as
\begin{align*}
\left\|Y_q-\widehat{Y}_q\right\|^2
&=\left\|Y_q\right\|^2-2\left\langle Y_q,\,Y_q\Pi_{\mathcal{U}}\right\rangle+\left\|Y_q\Pi_{\mathcal{U}}\right\|^2\\
&=\left\|Y_q\right\|^2-2\left\|Y_q\Pi_{\mathcal{U}}\right\|^2+\left\|Y_q\Pi_{\mathcal{U}}\right\|^2\\
&=\left\|Y_q\right\|^2-\left\|Y_q\Pi_{\mathcal{U}}\right\|^2,
\end{align*}
where the second equality holds since $\Pi_{\mathcal{U}}$ is idempotent and symmetric. Summing over the $k$ queries, we obtain
\[
\left\|Y-\widehat{Y}\right\|_F^2=\|Y\|_F^2-\left\|Y\Pi_{\mathcal{U}}\right\|_F^2 ,
\]
hence minimizing the reconstruction error is equivalent to maximizing the captured energy.

\textbf{Attribution.}
To relate the linear datamodeling score to the orthogonal projection, we first reduce the objective through a series of approximations. Fixing a query $q$, we can write
\begin{align*}
\mathrm{LDS}_q
&=\mathrm{Spearman}\text{-}\rho_S\left(-\langle\widehat{Y}_q,\mathbf{1}_{S}\rangle,\ f_q(\mathbf{1}_n-\mathbf{1}_{S})-f_q(\mathbf{1}_n)\right)\\
&\overset{\text{(1)}}{\appropto}\mathrm{Pearson}\text{-}\rho_S\left(-\langle\widehat{Y}_q,\mathbf{1}_{S}\rangle,\ f_q(\mathbf{1}_n-\mathbf{1}_{S})-f_q(\mathbf{1}_n)\right)\\
&\overset{\text{(2)}}{\approx}\mathrm{Pearson}\text{-}\rho_S \left(-\langle\widehat{Y}_q,\mathbf{1}_{S}\rangle,\ -\langle Y_q,\mathbf{1}_{S}\rangle\right)\\
&\overset{\text{(3)}}{=}\mathrm{Pearson}\text{-}\rho\left(\widehat{Y}_q,Y_q\right)\\
&\overset{\text{(4)}}{\approx}\ \frac{\left\langle Y_q\Pi_{\mathcal{U}},\,Y_q\right\rangle}{\left\|Y_q\Pi_{\mathcal{U}}\right\|\left\|Y_q\right\|}\\
&\overset{\text{(5)}}{=}\frac{\left\langle Y_q\Pi_{\mathcal{U}},\,Y_q\Pi_{\mathcal{U}}\right\rangle}{\left\|Y_q\Pi_{\mathcal{U}}\right\|\left\|Y_q\right\|}
=\frac{\left\|Y_q\Pi_{\mathcal{U}}\right\|^{2}}{\left\|Y_q\Pi_{\mathcal{U}}\right\|\left\|Y_q\right\|}
=\frac{\left\|Y_q\Pi_{\mathcal{U}}\right\|}{\left\|Y_q\right\|}.
\end{align*}
In (1), we replace Spearman correlation with Pearson correlation, since under approximate joint normality of the two sums the two quantities are related monotonically \citep{kruskal_ordinal_1958}, so improving one improves the other. In (2), we replace the ground truth by its MAGIC prediction. We defer the calculation of (3) to Appendix~\ref{appendix:proof_design}. The approximation in (4) is due to the empirical observation that the rows of \(Y\) are nearly mean-zero and centering has a negligible effect on the correlation. Finally, (5) holds since $\Pi_{\mathcal{U}}$ is idempotent and symmetric.

Summing over the $k$ queries and writing $D=\diag\left(\|Y_1\|,\ldots,\|Y_k\|\right)$ and $\widetilde{Y}=D^{-1}Y$,
\begin{equation}
\sum_{q=1}^k\mathrm{LDS}_q^2
\appropto\sum_{q=1}^k\frac{\left\|Y_q\Pi_{\mathcal{U}}\right\|^2}{\left\|Y_q\right\|^2}
=\left\|D^{-1}Y\Pi_{\mathcal{U}}\right\|_F^2
=\left\|\widetilde{Y}\Pi_{\mathcal{U}}\right\|_F^2 .
\label{eq:capture}
\end{equation}
Thus, we see that optimizing for the LDS likewise maximizes the captured energy of $\widetilde{Y}$.

\subsection{Selecting the Combinations to Measure}
\label{sec:method-selection}

The top-$B$ right-singular subspace of $Y$ is optimal for reconstruction, and that of $\widetilde{Y}$ is optimal for the attribution surrogate in \Eqref{eq:capture} \citep{fan_theorem_1949}. While the truncated SVD is ordinarily computed from the full matrix, it can also be assembled one direction at a time. To do so, we can probe (i.e., send through replay) the leading eigenvector of the residual Gram matrix, 
\[
R=Y\left(I_n-\Pi_{\mathcal{U}}\right)Y^\top,
\]
where we suppress the iteration index so that $\mathcal{U}$ and $R$ denote the current span and residual. After $B$ iterations, the span of the measurements recovers the top-$B$ right-singular subspace \citep{hotelling_analysis_1933, golub_matrix_2013}.

\begin{algorithm}[t]
\caption{\textsc{Mage} and \textsc{Spell}.}
\label{alg:mage-spell}
\begin{algorithmic}
\setlength{\baselineskip}{13pt}
\State \textbf{Input:} query gradients $V$, replay budget $B$, warm-up length $a_0$ (\textsc{Spell} only)
\State \textbf{Requires:} $\textsc{Replay}(s)=s^\top J$,
    $\textsc{Forward}(u)=Ju^\top$
\State $\widehat Y\gets0$, $\mathcal U\gets\{0\}$
\For{$a=1,\ldots,B$}
    \State $z_a\gets\Call{Mage-Select}{a}$
        or $\Call{Spell-Select}{a,\widehat Y}$
    \State $u\gets\Call{Replay}{Vz_a}$
        \Comment{One replay: $u=z_a^\top Y$}
    \State $c\gets V^\top\,\Call{Forward}{u}$
        \Comment{One forward-mode pass: $c=Yu^\top$}
    \State $\widehat Y\gets
        \Call{Extend-Projection}{\widehat Y,\mathcal U,u,c}$
    \State $\mathcal U\gets\mathcal U+\operatorname{span}(u)$
\EndFor
\State \Return $\widehat Y$ \Comment{$\widehat Y=Y\Pi_{\mathcal U}$}
\Statex
\Procedure{Mage-Select}{$a$}
    \State \Return the $a$th leading eigenvector of $G_V=V^\top V$
\EndProcedure
\Statex
\Procedure{Spell-Select}{$a,\widehat Y$}
    \State \textbf{if} $a\le a_0$ \textbf{then return}
        $\Call{Mage-Select}{a}$ \Comment{Warm up}
    \State $\widehat D\gets
        \diag(\max\{\|\widehat Y_q\|,\tau\})$
        \Comment{Estimate row norms}
    \State $G_{\mathrm{res}}\gets
        \Call{Residual-Gram}{V,[z_1,\ldots,z_{a-1}]}$
        \Comment{Project out replayed gradients}
    \State \Return $\widehat D^{-1}v_{\max}(\widehat D^{-1}G_{\mathrm{res}}\widehat D^{-1})$
    \Comment{$v_{\max}$: leading eigenvector}
\EndProcedure
\end{algorithmic}
\end{algorithm}

However, forming the residual Gram matrix would require the rows of $Y$ (or \(\widetilde Y\) analogously). Instead, \textsc{Mage} and \textsc{Spell} run the iterations using principled stand-ins built from the query gradients and the measurements taken so far, with \textsc{Mage} tracking the truncated SVD of $Y$ and \textsc{Spell} that of its row-normalized counterpart. 

\textbf{\textsc{Mage}.} \textsc{Mage} substitutes the query Gram matrix $G_V=V^\top V$ for $R$ and probes its leading eigenvectors in order. We also note that
\[
z^\top G_Vz=\|Vz\|^2
\]
measures the energy going into replay and $z^\top Rz$ the unmeasured energy after replay, so the substitution is reasonable whenever $J$ preserves the inner products between query gradients.

\textbf{\textsc{Spell}.}
To derive the measurements for \textsc{Spell}, we first note that for any probe \(r\) on \(\widetilde Y\), 
\[
r^\top\widetilde{Y}=(D^{-1}r)^\top Y,
\]
which is equivalent to probing $z\propto D^{-1}r$ on $Y$. Since the residual Gram matrix of \(\widetilde Y\) is 
\[
\widetilde{Y}\left(I_n-\Pi_{\mathcal{U}}\right)\widetilde{Y}^\top=D^{-1}RD^{-1},
\]
we can probe $v_{\max}(D^{-1}RD^{-1})$ on \(\widetilde Y\), which amounts to probing $z \propto D^{-1}v_{\max}(D^{-1}RD^{-1})$ on $Y$. Here, both $D$ and $R$ are unknown. We can first approximate the entries of \(D\) with
\[
\widehat{d}_q=\|\widehat{Y}_q\|,
\]
the length of the measured part of row \(q\). While \(\widehat d_q\) underestimates \(\|Y_q\|\) due to the unmeasured component, we note that every row is ``cut'' by the same subspace. Thus, the rankings across queries can stabilize after a few measurements. Since the row norm, and thus its estimate, enters the probe quadratically, any query whose row norm is badly underestimated could commandeer the next probe. To prevent this, we impose a percentile floor
\[
\widehat{D}=\diag\left(\max\{\widehat{d}_q,\tau\}\right),
\qquad
\tau=Q_{0.1}\left(\{\widehat{d}_q:\widehat{d}_q>0\}\right),
\]
with $Q_{0.1}$ the empirical tenth percentile. In practice, we observe that using nearby percentiles ($Q_{0.05}$, $Q_{0.15}$, or $Q_{0.2}$) does not noticeably change the estimate.

To approximate $R$, \textsc{Spell} follows \textsc{Mage}'s substitution of $G_V$. However, since the maximizer is now taken against the approximation\[\widehat{D}^{-1}G_V\widehat{D}^{-1} \approx D^{-1} RD^{-1},\] where \(\widehat{D}^{-1}\) is updated after every round, the probes no longer come from one fixed eigenbasis. Consequently, the deflation that \textsc{Mage} performs implicitly by probing in order, which removes directions already measured before each round, must now be carried out explicitly. 

Since the final estimate depends on any probe $z$ only through the combined query gradient $Vz$, it follows that deflating $z$ amounts to removing the part of $Vz$ measured by the previous replays, which we can achieve by solving the least squares problem
\[
\min_\beta\left\|Vz-VZ\beta\right\|,
\qquad
Z=[z_1,\ldots,z_{a-1}],
\]
with the minimizer given by $\beta=(Z^\top G_VZ)^{+}Z^\top G_Vz$. Writing the map $z\mapsto z-Z\beta$ in matrix form, we obtain the \textsc{Residual-Gram} matrix in \Algref{alg:mage-spell},
\[
P=I_k-Z\left(Z^\top G_VZ\right)^{+}Z^\top G_V,
\qquad
G_{\mathrm{res}}=P^\top G_VP.
\]
Assembling the approximation and residual projection, we arrive at
\[
z\;\propto\;\widehat{D}^{-1}v_{\max}\!\left(\widehat{D}^{-1}G_{\mathrm{res}}\widehat{D}^{-1}\right).
\]
Finally, since the estimates may be unreliable in the initial iterations, we warm up \textsc{Spell} by following \textsc{Mage} for the first $a_0$ rounds.

\section{Experiments}
\label{sec:experiments}
In this section, we evaluate \textsc{Mage} and \textsc{Spell} across two settings. In both settings, we first compute the exact influence matrix \(Y\) by running a separate replay for each of the \(k\) queries. For \textsc{Spell}, we set the first $a_0=\max(\lceil 0.1B\rceil,10)$ calls for the warm-up phase in \Algref{alg:mage-spell}.

We compare our methods against three baselines at the same budget of $B$ replays. Random probing replaces \textsc{Mage}'s eigendirection with a random Gaussian combination $z \sim \mathcal{N}(0, I_k)$ of the queries. The first-$B$-queries baseline replaces it with the coordinate vector $z = e_a$ for $a=1,\ldots,B$, where these \(B\) queries are chosen by class-balanced round robin on CIFAR-10 and by random permutation on TinyStories.

Principal component analysis (PCA) reconstructs \(\widehat Y = V_B^\top J\), where \(V_B\) is the rank-\(B\) SVD of \(V\), by passing the top-\(B\) principal directions of the query gradients \(V\) directly through replay. We also include two oracle baselines computable only with access to \(Y\), which are the rank-\(B\) singular value decomposition (SVD) of \(Y\), which achieves the smallest reconstruction error at rank \(B\), and the rank-\(B\) SVD of the row-normalized counterpart of \(Y\).

We evaluate all methods on the two aforementioned objectives in \Secref{subsec:objectives}. For reconstruction, we report the relative Frobenius error \[\left\|Y-\widehat{Y}\right\|_F\bigg/\Big\|Y\Big\|_F.\] For data attribution, we report the linear datamodeling score (LDS). To approximate this quantity for a given test query \(q\) and removal fraction \(p\), we take the following steps:
\begin{enumerate}
    \item Draw $M$ subsets $S_1,\ldots,S_M\subseteq[n]$ of size $np$ uniformly at random to be removed from the training set.
    \item Retrain the model on each subset \([n] \setminus S_j\) by setting the weight of every removed example to zero, holding all other training randomness fixed. 
    \item Record for every test query \(q\) the predicted change \(\widehat{\Delta}_q(S_j)=-\langle\widehat{Y}_q, \mathbf{1}_{S_j}\rangle\) and true change \(\Delta_q(S_j)=f_q(\mathbf{1}_n - \mathbf{1}_{S_j}) - f_q(\mathbf{1}_n)\) in the model output function.
    \item Compute the Spearman correlation between the predicted and true change over all \(M\) subsets.
\end{enumerate}
Finally, we average the LDS over all \(k\) test queries, which gives
\[\overline{\mathrm{LDS}} = \frac{1}{k}\sum_{q = 1}^{k} \rho\left(\Big[\widehat\Delta_q(S_j)\Big]_{j = 1}^{M}, \Big[\Delta_q(S_j)\Big]_{j = 1}^{M}\right).\]
We use $M=300$ retrained models at two removal fractions of $1\%$ and $5\%$, and hold the subsets fixed across test queries and methods. We also plot the estimate given by the exact influence matrix \(-\langle Y_q, \mathbf{1}_{S_j}\rangle \)
as the ceiling (MAGIC). 

\subsection{ResNet-9 on CIFAR-10}
\label{subsec:resnet9}

We first consider a ResNet-9 architecture trained on CIFAR-10 \citep{krizhevsky_learning_2009}, where our goal is to attribute the cross-entropy loss on up to \(k = 1000\) held-out test images back to the training data. The model contains 10M parameters and is trained on the full set of 50K training examples (see Appendix~\ref{appendix:resnet} for further details). We then average the results over ten random seeds, where the resulting models attain a mean $89.88\%$ test accuracy on the full CIFAR-10 test set. For the figures below, we plot the relative Frobenius error and linear datamodeling score against the number of replays \(B\) over the total number of test queries \(k\).

\begin{figure}[tbh!]
\centering
\includegraphics[width=\linewidth]{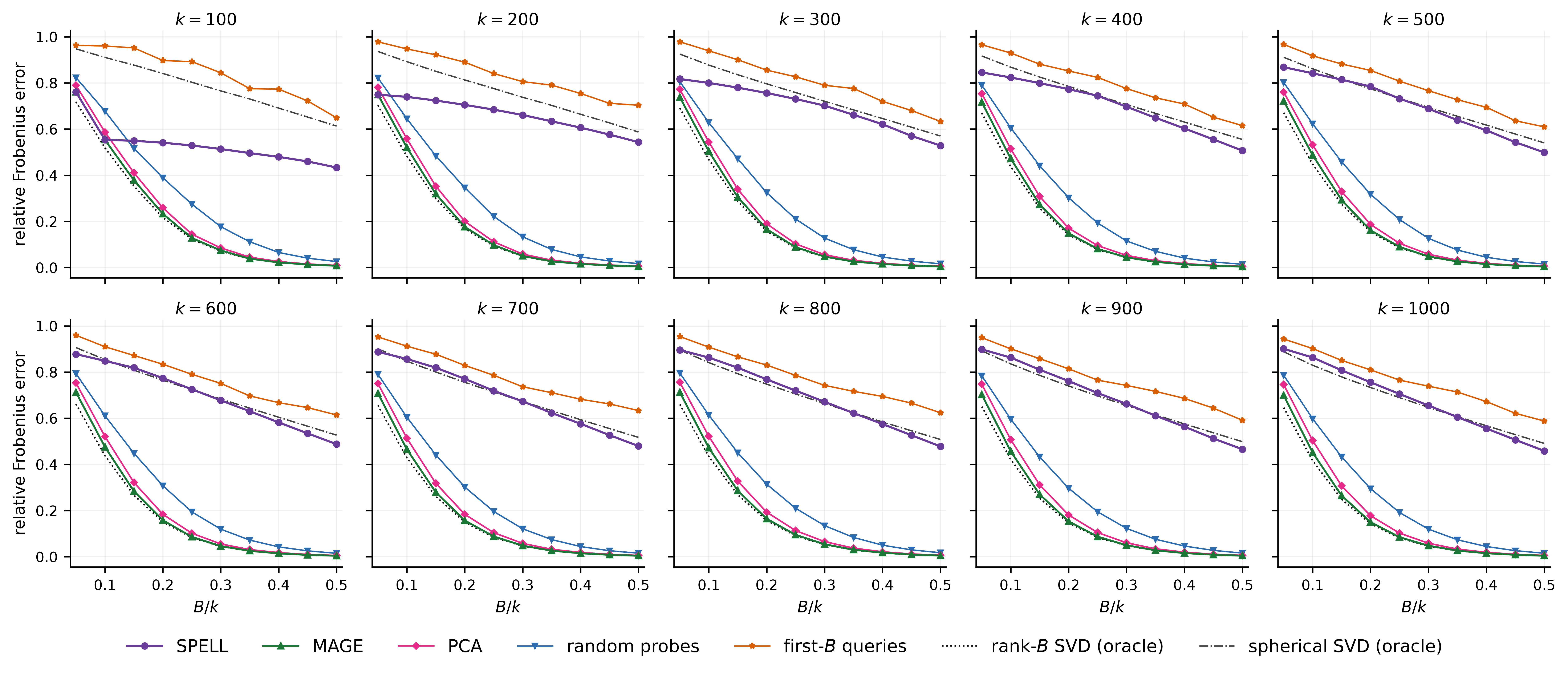}
\caption{Reconstruction error across 10 seeded ResNet-9 models trained on CIFAR-10 and query set size \(k\). \textsc{Mage} is nearly indistinguishable from the oracle rank-$B$ SVD.}
\label{fig:cifar-grid-recon}
\end{figure}

\begin{figure}[tbh!]
\centering
\includegraphics[width=\linewidth]{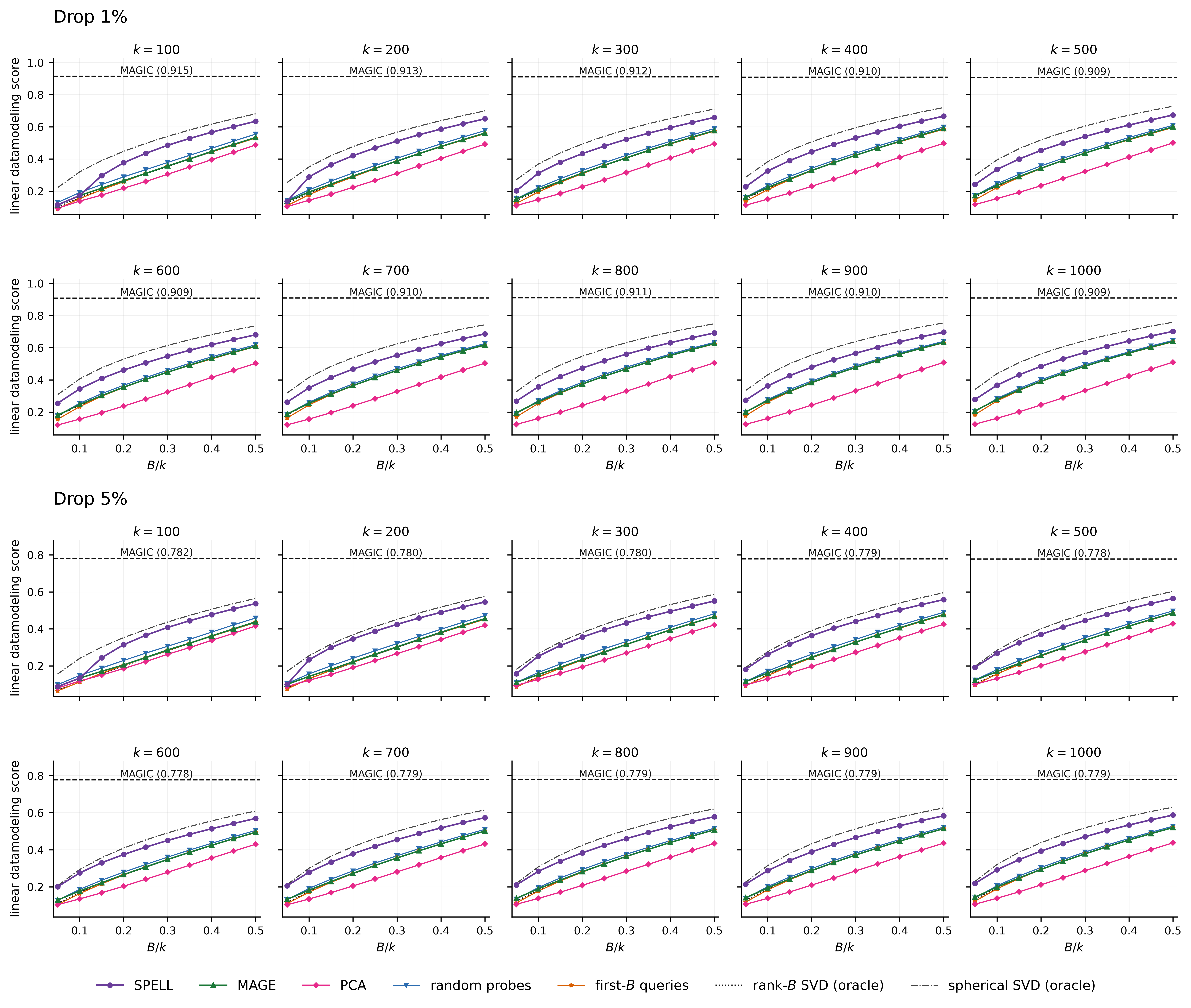}
\caption{Linear datamodeling score across 10 seeded ResNet-9 models trained on CIFAR-10 and query set size \(k\) after removing a random $1\%$ (top) and 5\% (bottom) of the training set. \textsc{Spell} outperforms all computable methods at every \(k\) once \(B \ge 0.2k\).}
\label{fig:cifar-grid-lds}
\end{figure}

We see in \Figref{fig:cifar-grid-recon} that \textsc{Mage} attains the lowest reconstruction error of any implementable method at every budget and query set size. For data attribution, the top panel of \Figref{fig:cifar-grid-lds} shows that \textsc{Spell} consistently outperforms computable alternatives at every size \(k\) once the budget reaches a fifth of the query set size, when 1\% of the training data is dropped. Furthermore, as shown in the bottom panel of \Figref{fig:cifar-grid-lds}, \textsc{Spell} retains the same advantage when 5\% of the training data is dropped. 

Table~\ref{tab:cifar-summary} details both metrics at $k = 500$ under the budgets of $B = 50$ and $B = 100$. At $B = 50$, \textsc{Mage} attains a relative Frobenius error of $0.49$ against $0.53$ for PCA, $0.62$ for random probes, and $0.92$ for the first-$B$ queries, while \textsc{Spell} obtains an LDS of $0.34$ against $0.25$ for random probes, $0.22$ for the first-$B$ queries, and $0.15$ for PCA when dropping 1\% of the data.

\begin{figure}[tbh!]
\centering
\includegraphics[width=\linewidth]{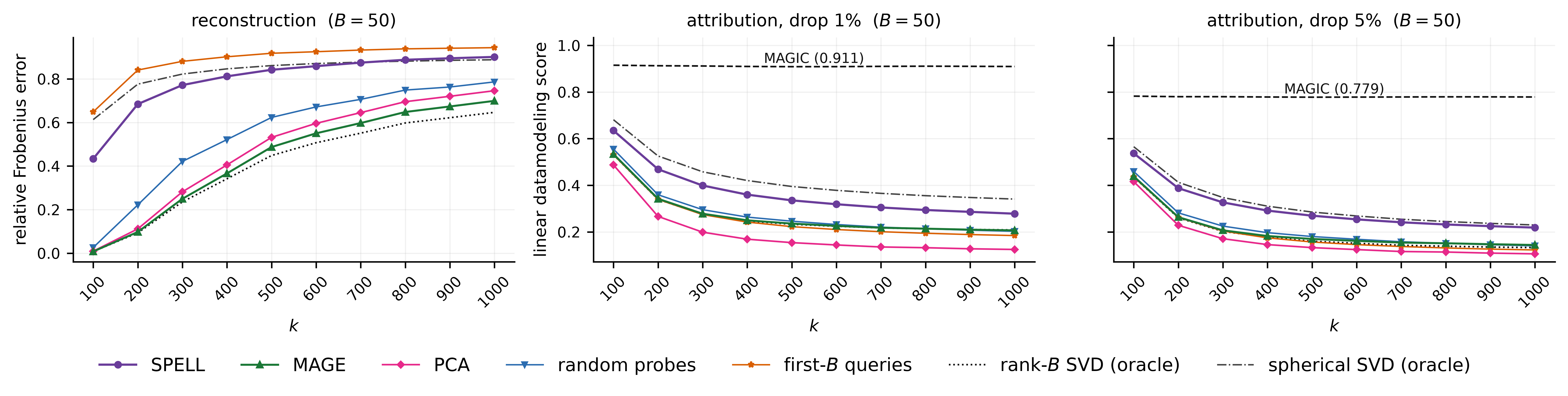}
\caption{At a fixed budget of replays \(B = 50\), we see that reconstruction error increases and attribution quality decays as the size \(k\) of the test query set grows.}
\label{fig:cifar-law}
\end{figure}

Finally, in \Figref{fig:cifar-law}, we fix the number of replays at \(B = 50\) and vary the size \(k\) of the test query set. We see that both the relative Frobenius error and linear datamodeling score across all methods degrade as \(k\) increases. Nevertheless, both \textsc{Mage} and \textsc{Spell} degrade gracefully and retain their comparative advantages over other methods.

\begin{table}[tbh!]
\centering
\small
\begin{tabular}{@{}lc@{\hspace{4pt}}lRRR@{}}
\toprule
 & & & \multicolumn{1}{c}{} & \multicolumn{2}{c}{Linear datamodeling score $\uparrow$} \\
\cmidrule(l){5-6}
Budget & & Method & \multicolumn{1}{c}{Relative Frobenius error $\downarrow$} & \multicolumn{1}{c}{$1\%$ removed} & \multicolumn{1}{c}{$5\%$ removed} \\
\midrule
\multirow{7}{*}{$B = 50$} & \kMage & \textsc{Mage} & \best{0.4872}\se{0.0064} & 0.2358\se{0.0017} & 0.1698\se{0.0032} \\
 & \kSpell & \textsc{Spell} & 0.8419\se{0.0061} & \best{0.3353}\se{0.0023} & \best{0.2696}\se{0.0045} \\
 & \kRandom & Random probes & 0.6231\se{0.0049} & 0.2461\se{0.0013} & 0.1803\se{0.0032} \\
 & \kPCA & PCA & 0.5316\se{0.0061} & 0.1541\se{0.0010} & 0.1327\se{0.0017} \\
 & \kFirst & First $B$ queries & 0.9177\se{0.0040} & 0.2226\se{0.0024} & 0.1568\se{0.0032} \\
\cmidrule(l){2-6}
 & \kSVD & Rank-$B$ SVD & 0.4483\se{0.0058} & 0.2328\se{0.0022} & 0.1599\se{0.0030} \\
 & \kSphSVD & Spherical SVD & 0.8611\se{0.0049} & 0.3948\se{0.0023} & 0.2852\se{0.0056} \\
\midrule
\multirow{7}{*}{$B = 100$} & \kMage & \textsc{Mage} & \best{0.1614}\se{0.0034} & 0.3425\se{0.0017} & 0.2570\se{0.0040} \\
 & \kSpell & \textsc{Spell} & 0.7845\se{0.0064} & \best{0.4538}\se{0.0023} & \best{0.3708}\se{0.0052} \\
 & \kRandom & Random probes & 0.3177\se{0.0045} & 0.3572\se{0.0013} & 0.2721\se{0.0041} \\
 & \kPCA & PCA & 0.1869\se{0.0036} & 0.2337\se{0.0013} & 0.2006\se{0.0021} \\
 & \kFirst & First $B$ queries & 0.8542\se{0.0048} & 0.3405\se{0.0020} & 0.2538\se{0.0034} \\
\cmidrule(l){2-6}
 & \kSVD & Rank-$B$ SVD & 0.1545\se{0.0031} & 0.3420\se{0.0016} & 0.2545\se{0.0040} \\
 & \kSphSVD & Spherical SVD & 0.7745\se{0.0047} & 0.5189\se{0.0026} & 0.4007\se{0.0057} \\
\midrule
$B = k$ & \kMagic & MAGIC & \multicolumn{1}{c}{--} & 0.9087\se{0.0038} & 0.7778\se{0.0058} \\
\bottomrule
\end{tabular}
\caption{Performance of methods on ResNet-9 at $k = 500$ with $B = 50$ and $100$ on CIFAR-10, evaluated against models retrained after removing a random $1\%$ and $5\%$ of the training set. We \textbf{bold} the best computable method in each column, where MAGIC is the ceiling given by the exact influence matrix at $B = k$. The values are averaged over ten runs and reported along with their standard errors.}
\label{tab:cifar-summary}
\end{table}

\newpage
\subsection{GPT-2 on TinyStories}
\label{subsec:gpt2}

For our second domain, we fine-tune a 12-layer GPT-2-style decoder \citep{radford_language_2019} on TinyStories \citep{eldan_tinystories_2023}. We consider a chunk of $512$ consecutive tokens as a single training example, and the query value is the negative log-likelihood the model assigns to the single token following a held-out chunk of the same length. We fine-tune the model on \(2048\) training examples, where our goal is to attribute up to \(600\) test queries back to the training data. The model has 92M parameters, and only the fine-tuning stage is attributed starting from a fixed pretrained checkpoint shared across all runs (see Appendix~\ref{appendix:lm} for further details). We then average the results over four runs, where the fine-tuned models attain a mean $65.04\%$ top-1 next-token accuracy on the held-out query corpus.

\begin{figure}[tbh!]
\centering
\includegraphics[width=\linewidth]{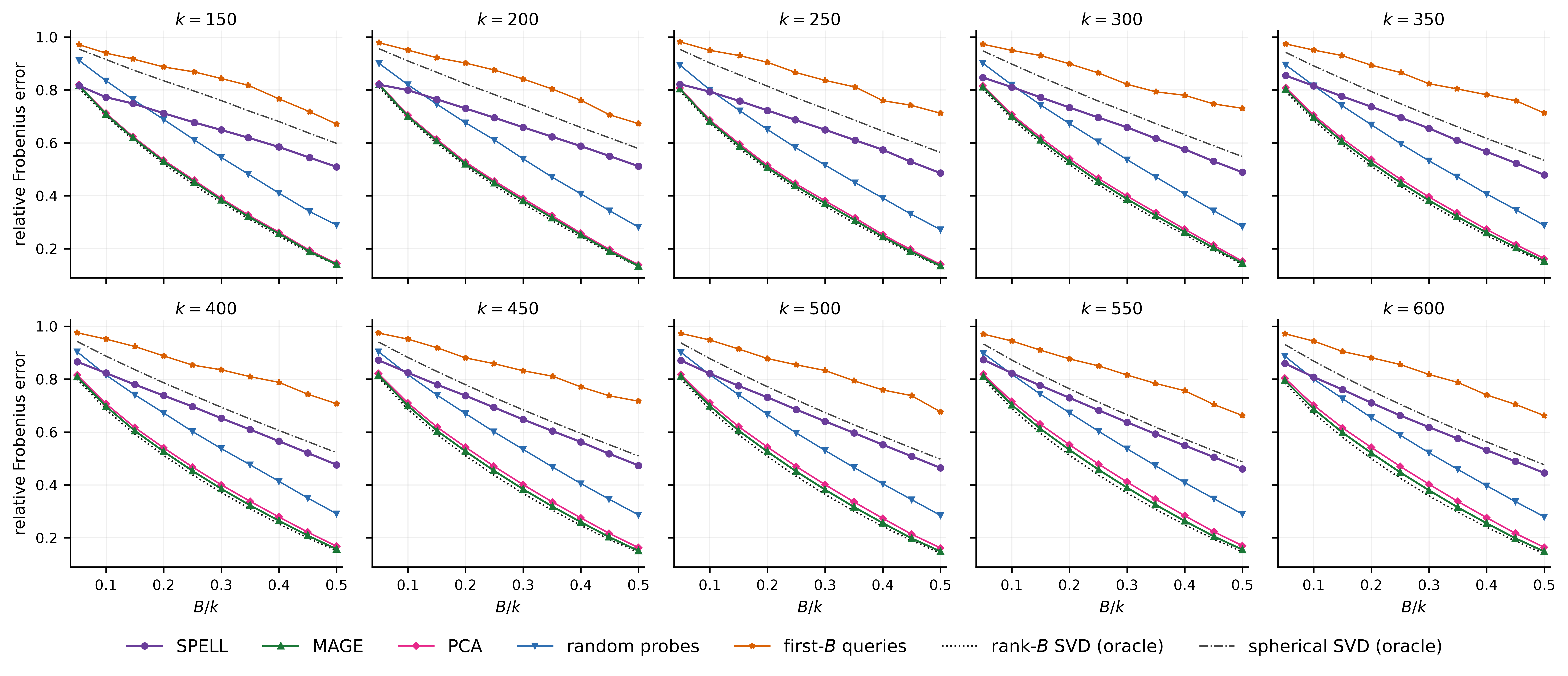}
\caption{Reconstruction error across 4 seeded GPT-2-style decoders fine-tuned on TinyStories and query set
size \(k\). \textsc{Mage} is nearly indistinguishable from the oracle
rank-$B$ SVD.}
\label{fig:lm-grid-recon}
\end{figure}

\begin{figure}[tbh!]
\centering
\includegraphics[width=\linewidth]{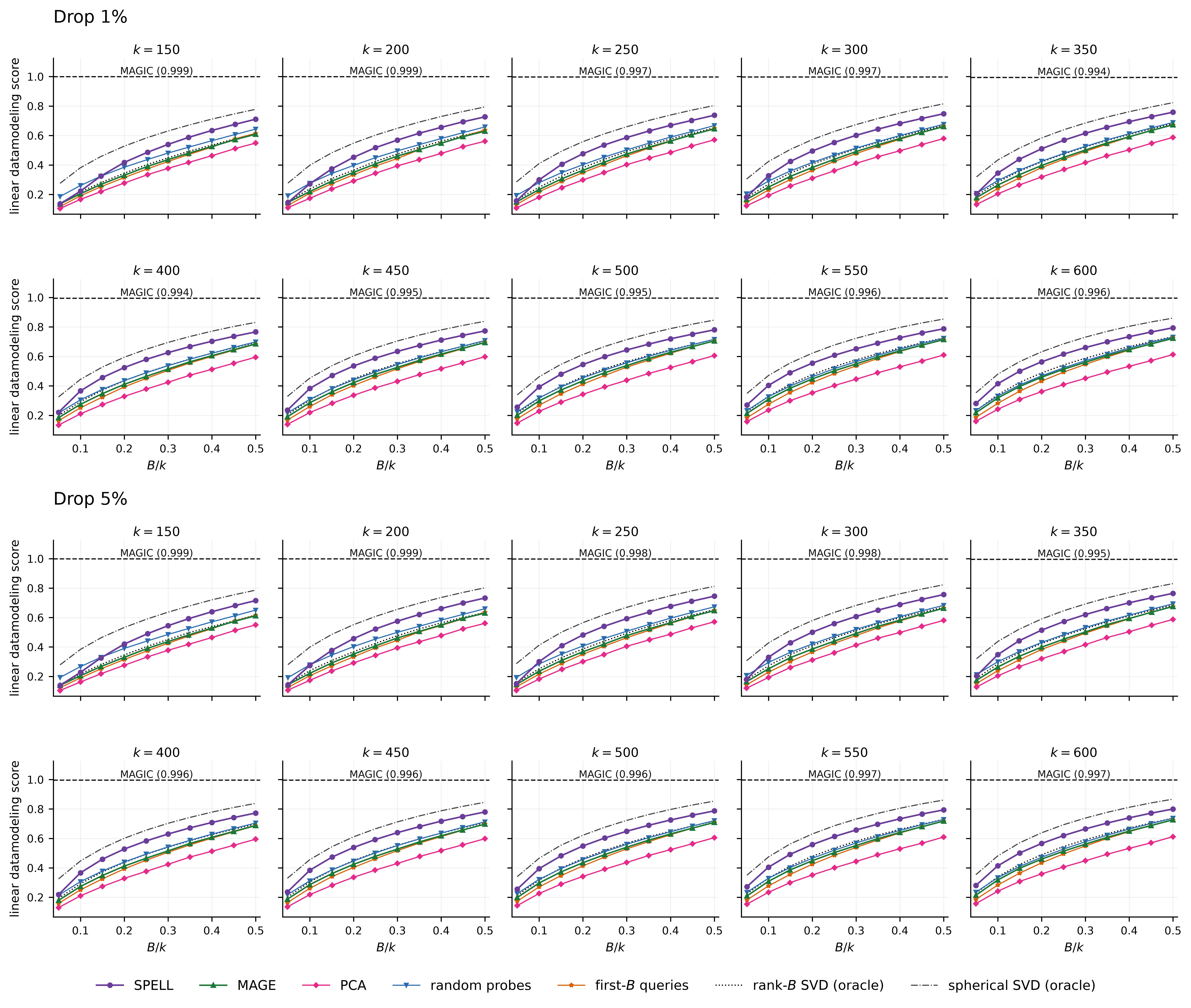}
\caption{Linear datamodeling score across 4 seeded GPT-2-style decoders fine-tuned on TinyStories and query set size \(k\) after removing a random $1\%$ (top) and 5\% (bottom) of the fine-tuning set. \textsc{Spell} outperforms all computable methods at every $k$ once \(B \ge 0.2k\).}
\label{fig:lm-grid-lds}
\end{figure}

\begin{table}[tbh!]
\centering
\small
\begin{tabular}{@{}lc@{\hspace{4pt}}lRRR@{}}
\toprule
 & & & \multicolumn{1}{c}{} & \multicolumn{2}{c}{Linear datamodeling score $\uparrow$} \\
\cmidrule(l){5-6}
Budget & & Method & \multicolumn{1}{c}{Relative Frobenius error $\downarrow$} & \multicolumn{1}{c}{$1\%$ removed} & \multicolumn{1}{c}{$5\%$ removed} \\
\midrule
\multirow{7}{*}{$B = 50$} & \kMage & \textsc{Mage} & \best{0.6986}\se{0.0012} & 0.2980\se{0.0018} & 0.2970\se{0.0017} \\
 & \kSpell & \textsc{Spell} & 0.8206\se{0.0013} & \best{0.3931}\se{0.0004} & \best{0.3947}\se{0.0017} \\
 & \kRandom & Random probes & 0.8165\se{0.0010} & 0.3191\se{0.0014} & 0.3221\se{0.0018} \\
 & \kPCA & PCA & 0.7104\se{0.0012} & 0.2275\se{0.0022} & 0.2258\se{0.0012} \\
 & \kFirst & First $B$ queries & 0.9478\se{0.0003} & 0.2683\se{0.0015} & 0.2704\se{0.0014} \\
\cmidrule(l){2-6}
 & \kSVD & Rank-$B$ SVD & 0.6847\se{0.0009} & 0.3164\se{0.0027} & 0.3168\se{0.0023} \\
 & \kSphSVD & Spherical SVD & 0.8777\se{0.0008} & 0.4618\se{0.0016} & 0.4654\se{0.0018} \\
\midrule
\multirow{7}{*}{$B = 100$} & \kMage & \textsc{Mage} & \best{0.5257}\se{0.0013} & 0.4341\se{0.0009} & 0.4350\se{0.0017} \\
 & \kSpell & \textsc{Spell} & 0.7310\se{0.0012} & \best{0.5451}\se{0.0006} & \best{0.5488}\se{0.0009} \\
 & \kRandom & Random probes & 0.6665\se{0.0009} & 0.4543\se{0.0005} & 0.4566\se{0.0021} \\
 & \kPCA & PCA & 0.5429\se{0.0013} & 0.3428\se{0.0009} & 0.3416\se{0.0012} \\
 & \kFirst & First $B$ queries & 0.8777\se{0.0002} & 0.4120\se{0.0010} & 0.4136\se{0.0011} \\
\cmidrule(l){2-6}
 & \kSVD & Rank-$B$ SVD & 0.5075\se{0.0012} & 0.4622\se{0.0006} & 0.4635\se{0.0015} \\
 & \kSphSVD & Spherical SVD & 0.7711\se{0.0004} & 0.6130\se{0.0007} & 0.6217\se{0.0018} \\
\midrule
$B = k$ & \kMagic & MAGIC & \multicolumn{1}{c}{--} & 0.9952\se{0.0004} & 0.9964\se{0.0002} \\
\bottomrule
\end{tabular}
\caption{Performance of methods on GPT-2 at $k = 500$ with $B = 50$ and $100$ on TinyStories, evaluated against models retrained after removing a random $1\%$ and $5\%$ of the fine-tuning set. We \textbf{bold} the best computable method in each column, and MAGIC is the ceiling given by the exact influence matrix at $B = k$. The values are averaged over four runs and reported along with their standard errors.}
\label{tab:lm-summary}
\end{table}

In \Figref{fig:lm-grid-recon}, we see that \textsc{Mage} attains the lowest reconstruction error of any implementable method at every budget and query set size. For data attribution, the top panel of \Figref{fig:lm-grid-lds} shows that \textsc{Spell} also outperforms every computable alternative at every \(k\) once the budget reaches a fifth of the query set size, when $1\%$ of the fine-tuning data is dropped. We also see in the bottom panel of \Figref{fig:lm-grid-lds} that \textsc{Spell} remains performant when $5\%$ of the data is dropped. In greater detail, Table~\ref{tab:lm-summary} shows that at $k = 500$ with budget $B = 50$, \textsc{Mage} attains a relative Frobenius error of $0.70$ against $0.71$ for PCA, $0.82$ for random probes, and $0.95$ for the first-$B$ queries, while \textsc{Spell} obtains an LDS of $0.39$ against $0.32$ for random probes, $0.27$ for the first-$B$ queries, and $0.23$ for PCA when dropping 1\% of the data.

\begin{figure}[tbh!]
\centering
\includegraphics[width=1.0\linewidth]{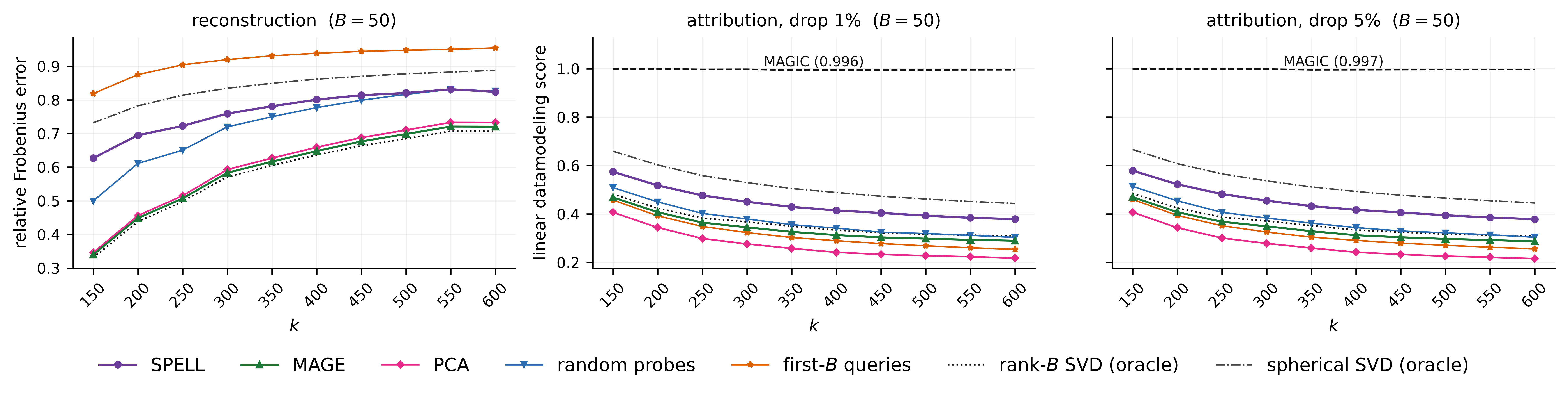}
\caption{At a fixed budget of replays \(B = 50\), reconstruction error increases and attribution quality decays as the size \(k\) of the test query set grows.}
\label{fig:lm-law}
\end{figure}

Finally, in \Figref{fig:lm-law}, we fix the budget at \(B = 50\) and vary the size \(k\) of the test query set, and observe the same degradation as $k$ increases. Nonetheless, both \textsc{Mage} and \textsc{Spell} maintain their comparative advantages over other methods. While the two scenarios are not directly comparable given that they contain distinct datasets, optimizers, and network architectures, we note that they nevertheless behave alike. Specifically, at the fixed budget $B = 50$, \textsc{Spell} attains $37\%$ of MAGIC's ceiling on CIFAR-10 and $39\%$ on TinyStories when dropping 1\% of the data, and $85\%$ of the performance of rank-\(B\) spherical SVD in both. The ranking among the competitive methods also does not differ across the two domains once \(B \ge 0.2k\).

\subsection{Reconciling the Two Objectives}

Across both settings, \textsc{Mage} attains the lowest reconstruction error whereas \textsc{Spell} attains the highest linear datamodeling score, which may seem
conflicting given that \textsc{Mage} recovers the influence matrix almost exactly in terms of the relative Frobenius error. We resolve this apparent contradiction by further examining the geometry of $Y$ in Appendix~\ref{appendix:geometry}.

\section{Conclusion}

In this paper, we cast data attribution at scale as a problem of estimating the influence matrix from a small number of measurements, using the observation that a single replay measures any linear combination of its rows. Next, we introduce \textsc{Mage} and \textsc{Spell}, algorithms tailored to recover the matrix itself and to preserve data attribution fidelity respectively. Across a ResNet-9 on CIFAR-10 and a GPT-2-style decoder on TinyStories, both methods consistently outperform every computable alternative on their targeted objectives. Given that the cost of both procedures is set by the measurement budget rather than by the number of queries, they enable efficient attribution for many queries simultaneously where recovering the exact influence matrix may be computationally prohibitive. 

One promising avenue for future work is to investigate whether better selection rules exist, and in particular whether the attribution objective admits a measurement that is optimal in the same way that truncated SVD is optimal for reconstruction, which would help guide the design of practical selection rules at a fixed measurement budget.

\section*{Acknowledgements}
We thank Sam Gunn and Nathan Ju for helpful discussions.

Y.C. and A.I. gratefully acknowledge support from the Delta system at the National Center for Supercomputing Applications through allocation CIS251020 from the Advanced Cyberinfrastructure Coordination Ecosystem: Services \& Support (ACCESS) program, which is supported by National Science Foundation grants \#2138259, \#2138286, \#2138307, \#2137603, and \#2138296. A.M. gratefully acknowledges support from National Science Foundation grant DMS-2515716 and a Google Faculty Award.

\bibliographystyle{unsrtnat}
\bibliography{references}

\appendix

\newpage

\section{Proof of the Pearson Correlation Identity}
\label{appendix:proof_design}

First, we define
\[H = I_n - \frac{1}{n}\mathbf{1}_n \mathbf{1}_n^\top, \qquad \alpha = \frac{p(1-p)n}{n-1}.\]
For a uniformly random subset \(S\) of size \(np\), \(\textrm{Cov}_S(\mathbf{1}_S) = \alpha H\), and we can write
\begin{align*}
    \mathrm{Pearson}\text{-}\rho_S\left(-\langle\widehat Y_q,\mathbf{1}_S\rangle,\ -\langle Y_q,\mathbf{1}_S\rangle\right)
    &= \frac{\widehat Y_q \mathrm{Cov}_S(\mathbf{1}_S)Y_q^\top}{\sqrt{\widehat Y_q \mathrm{Cov}_S(\mathbf 1_S) \widehat Y_q^\top} \sqrt{Y_q \mathrm{Cov}_S(\mathbf{1}_S) Y_q^\top}}\\
    &= \frac{\alpha \widehat Y_q H Y_q^\top}{\sqrt{\alpha \widehat Y_q H \widehat Y_q^\top} \sqrt{\alpha Y_q H Y_q^\top}}\\
    &= \frac{\widehat Y_q H Y_q^\top}{\sqrt{ \widehat Y_q H \widehat Y_q^\top} \sqrt{ Y_q H Y_q^\top}}\\
    &= \frac{\big(\widehat Y_q H\big) \big(Y_q H\big)^\top}{\big\|\widehat Y_q H\big\| \big\|Y_q H\big\|}\\
    &=\mathrm{Pearson}\text{-}\rho(\widehat Y_q, Y_q),
\end{align*}
where the penultimate equality holds because \(H\) is idempotent and symmetric.

\section{Architecture Details}

\subsection{ResNet-9 on CIFAR-10}
\label{appendix:resnet}

Our ResNet-9 consists of eight $3\times3$ convolutions, each followed by batch normalization and GELU nonlinearity, and a final linear head. The first layer widens the input image to 80 channels at full resolution, and the second layer widens it to 160 and applies a $2\times2$ average pooling. Two more layers at 160 channels form the first residual block. The next two layers widen and downsample twice more, to 320 channels at $8\times8$ and then to 640 channels at $4\times4$. The last two convolutional layers form a second residual block at 640 channels. A final $4\times4$ average pooling reduces each channel to a single value, and a linear head maps the resulting 640-dimensional vector to the ten class logits.

\begin{table}[tbh!]
\centering
\small
\begin{tabular}{lr}
\toprule
Hyperparameter & Value \\
\midrule
Learning rate           & $0.085$ \\
Weight decay            & $5\times10^{-4}$ \\
Batch size              & $500$ \\
Epochs                  & $20$ \\
Training steps $T$      & $2{,}000$ \\
Training examples $n$   & $50{,}000$ \\
Optimizer               & SGD \\
Momentum                & $0.85$ \\
LR schedule             & One-cycle linear \\
LR start multiplier     & $0.1$ \\
LR end multiplier       & $0.0$ \\
LR peak time            & $0.3$ \\
\bottomrule
\end{tabular}
\caption{Training hyperparameters for ResNet-9 on CIFAR-10. The learning rate and the weight decay are both divided by the momentum correction $1+1/(1-0.85)$.}
\label{tab:resnet-hparams}
\end{table}

The $k=1000$ queries are randomly chosen to ensure balance at $100$
images per class and prefix-nested in $k$, so that the $k=100$ set is the first $100$ rows of the $k=1000$ set. 

\subsection{GPT-2 on TinyStories}
\label{appendix:lm}

Our model is a 12-layer pre-norm transformer
decoder of width $768$ with $12$ attention heads of $64$ dimensions. The input is a chunk of $512$ tokens and is mapped through an embedding table and added to a learned position embedding, where we tokenize with a vocabulary of $8192$ entries that contains the $7936$ most frequent words of the pretraining text together with $256$ single-byte fallbacks. Each block applies causal self-attention and a feed-forward network that widens from $768$ to $3072$ through a GELU nonlinearity, preceded by layer normalization and wrapped in a residual connection. 

We partition the TinyStories training corpus into chunks of length $512$ and divide them into three splits with no overlap, reserving $896{,}000$ chunks for pretraining, $4473$ for fine-tuning, and $1024$ for testing. We attribute on the test split, where each of our four runs fine-tunes on a different draw of $n = 2048$ chunks from the fine-tuning split.

\begin{table}[tbh!]
\centering
\small
\begin{tabular}{lrr}
\toprule
Hyperparameter & Pretraining & Fine-tuning \\
\midrule
Learning rate           & $6\times10^{-4}$ & $5\times10^{-6}$ \\
Weight decay            & $0.1$            & $10^{-5}$ \\
Batch size              & $128$            & $16$ \\
Epochs                  & $2$              & $2$ \\
Training steps $T$      & $14{,}000$       & $256$ \\
Training examples $n$   & $896{,}000$      & $2{,}048$ \\
Optimizer               & AdamW            & AdamW \\
$(\beta_1,\beta_2)$     & $(0.95, 0.999)$  & $(0.95, 0.999)$ \\
LR schedule             & One-cycle linear & One-cycle linear \\
LR start multiplier     & $10^{-6}$        & $10^{-6}$ \\
LR end multiplier       & $0.1$            & $0.1$ \\
LR peak time            & $0.25$           & $0.25$ \\
Matmul precision        & TF32             & fp32 \\
\bottomrule
\end{tabular}
\caption{Training hyperparameters for the language model on TinyStories.}
\label{tab:lm-hparams}
\end{table}

\section{Geometry of the Influence Matrix}
\label{appendix:geometry}

In \Secref{subsec:reduction}, we have shown that reconstruction and attribution weight queries differently, and have seen in \Secref{sec:experiments} that \textsc{Mage} and \textsc{Spell} indeed excel at their respective tasks. Here, we provide some further intuition behind this phenomenon. In particular, we show that for the two scenarios considered in \Secref{sec:experiments}, the rows of the influence matrix differ by many orders of magnitude in norm. 

\begin{figure}[tbh!]
\centering
\includegraphics[width=\linewidth]{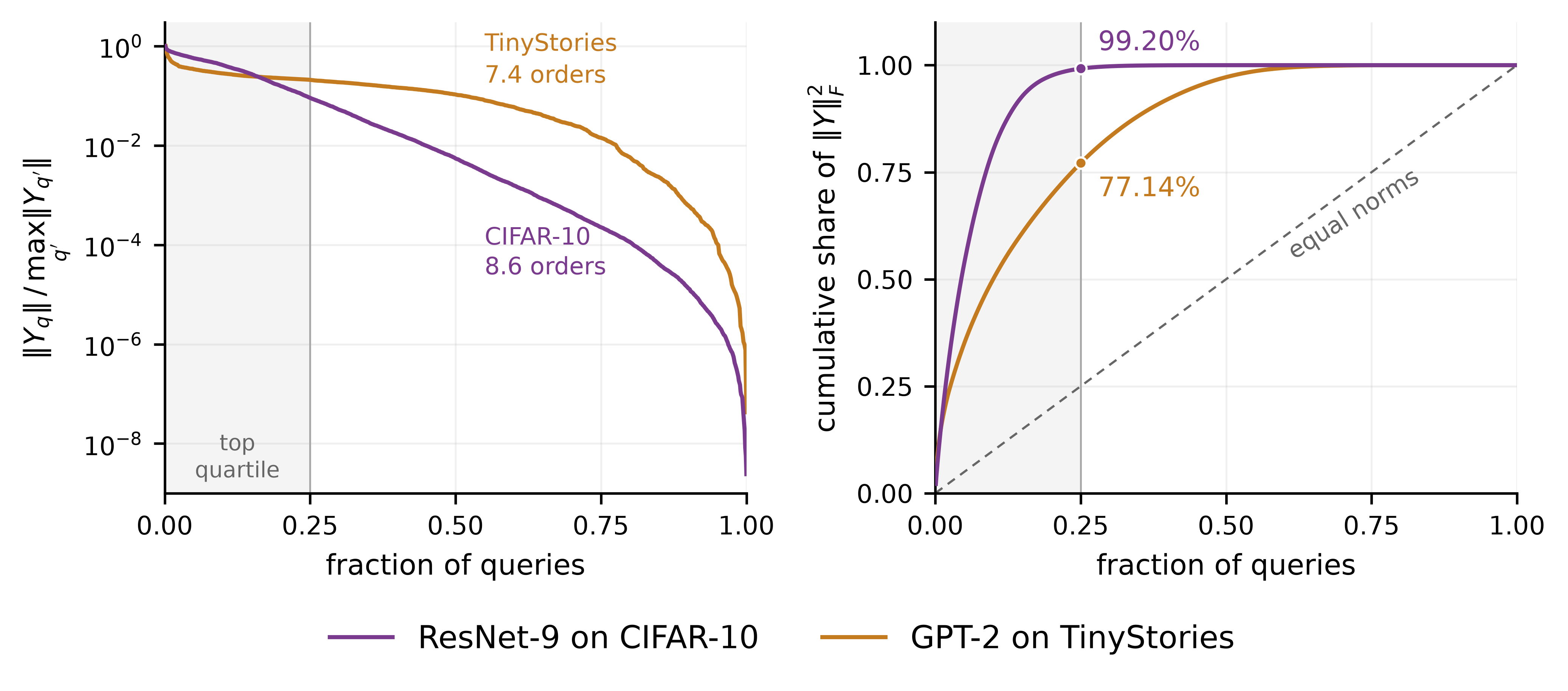}
\caption{Row norms of the exact influence matrix \(Y\) at $k = 1000$ on CIFAR-10 and $k = 600$ on TinyStories, sorted in decreasing order. The left panel plots the row norm of each query divided by the largest, and the right panel plots the fraction of the total energy the queries accumulate, where the shaded region represents the top quartile of queries. The values are averaged over ten runs for CIFAR-10 and four on TinyStories.}
\label{fig:geometry}
\end{figure}

In \Figref{fig:geometry}, we sort the queries using their row norms $\|Y_q\|$ in decreasing order and plot them along with the share of total energy accumulated thus far. For the ResNet-9 setting considered in \Secref{subsec:resnet9} with \(k = 1000\), the norms span $8.6$ orders of magnitude on CIFAR-10, and the largest quartile of queries holds $99.20\%$ of the total energy. For the GPT-2 setting shown in \Secref{subsec:gpt2} with \(k = 600\), the row norms span $7.4$ orders of magnitude, where the largest quartile of queries captures $77.14\%$ of the total energy on TinyStories.

Since the relative Frobenius error weights each query by the square of its row norm \(\|Y_q\|^2\), it follows that on CIFAR-10 the error is determined almost entirely by the top quartile. The linear datamodeling score instead averages over every query equally regardless of its norm. Consequently, the same estimate can be scored very differently under the two objectives. To measure this behavior, we consider the mean per-query relative error 
\[\frac{1}{k}\sum_{q = 1}^{k}\left\|Y_q-\widehat{Y}_q\right\|\Big/\Big\|Y_q\Big\|,\]
which normalizes each error by its ground-truth influence. Table~\ref{tab:decouple-global} shows that \textsc{Mage} attains a relative Frobenius error of $0.0045$ on CIFAR-10 at $k = 1000$ and $B = 500$ against a mean per-query relative error of $0.447$. In contrast, \textsc{Spell} attains a balanced performance of $0.458$ against $0.584$.

\begin{table}[tbh!]
\centering
\small
\begin{tabular}{llrrr}
\toprule
Setting & Method & Relative Frobenius error & Mean per-query error & Ratio \\
\midrule
\multirow{4}{*}{CIFAR-10}
 & \textsc{Mage}  & $0.0045$ & $0.447$ & $100$ \\
 & \textsc{Spell} & $0.458$  & $0.584$ & $1.3$ \\
 & Random probes  & $0.0152$ & $0.491$ & $32$ \\
 & PCA            & $0.0057$ & $0.576$ & $101$ \\
\midrule
\multirow{4}{*}{TinyStories}
 & \textsc{Mage}  & $0.148$  & $0.459$ & $3.1$ \\
 & \textsc{Spell} & $0.445$  & $0.545$ & $1.2$ \\
 & Random probes  & $0.279$  & $0.568$ & $2.0$ \\
 & PCA            & $0.164$  & $0.507$ & $3.1$ \\
\bottomrule
\end{tabular}
\caption{Performance of methods across ResNet-9 at $k = 1000$ with $B = 500$ on CIFAR-10, and GPT-2 at $k = 600$ with $B = 300$ on TinyStories. The values are averaged over ten runs for CIFAR-10 and four on TinyStories.}
\label{tab:decouple-global}
\end{table}

To capture a clearer picture of the underlying geometry, we sort the queries by $\|Y_q\|$ and partition them into quartiles. Table~\ref{tab:decouple-quartile} below reports the per-query relative error and the linear datamodeling score within every quartile. Across both scenarios, \textsc{Mage} reconstructs the largest quartile almost exactly but learns little about the smallest quartile, and its attribution quality follows the same pattern. In contrast, \textsc{Spell} remains nearly flat on both measures. 

While the smallest quartile may appear unpredictable, the exact influence matrix suggests otherwise, scoring $0.88$ on the smallest quartile and $0.92$ on the largest, where \textsc{Mage} attains only $0.35$ on the smallest against $0.59$ for \textsc{Spell} on CIFAR-10. This suggests that it could be helpful for practitioners to understand the imbalance in the row norms before deciding on the optimal attribution strategy. 

\begin{table}[tbh!]
\centering
\footnotesize
\setlength{\tabcolsep}{3.2pt}
\begin{tabular}{llrrrrcrrrrr}
\toprule
& & \multicolumn{4}{c}{Per-query relative error} && \multicolumn{5}{c}{Linear datamodeling score} \\
\cmidrule{3-6}\cmidrule{8-12}
Setting & Quartile & \textsc{Mage} & \textsc{Spell} & Random & PCA && \textsc{Mage} & \textsc{Spell} & Random & PCA & MAGIC \\
\midrule
\multirow{4}{*}{CIFAR-10}
 & Q1            & $0.88$ & $0.69$ & $0.88$ & $1.14$ && $0.35$ & $0.59$ & $0.35$ & $0.09$ & $0.88$ \\
 & Q2            & $0.87$ & $0.57$ & $0.82$ & $1.12$ && $0.37$ & $0.73$ & $0.44$ & $0.12$ & $0.92$ \\
 & Q3            & $0.04$ & $0.56$ & $0.25$ & $0.05$ && $0.91$ & $0.73$ & $0.87$ & $0.91$ & $0.92$ \\
 & Q4            & $0.00$ & $0.52$ & $0.02$ & $0.00$ && $0.92$ & $0.76$ & $0.92$ & $0.92$ & $0.92$ \\
\midrule
\multirow{4}{*}{TinyStories}
 & Q1            & $0.89$ & $0.64$ & $0.89$ & $0.98$ && $0.43$ & $0.69$ & $0.43$ & $0.20$ & $0.98$ \\
 & Q2            & $0.85$ & $0.54$ & $0.73$ & $0.94$ && $0.48$ & $0.81$ & $0.63$ & $0.28$ & $1.00$ \\
 & Q3            & $0.09$ & $0.54$ & $0.41$ & $0.10$ && $0.98$ & $0.82$ & $0.90$ & $0.97$ & $1.00$ \\
 & Q4            & $0.01$ & $0.45$ & $0.24$ & $0.01$ && $1.00$ & $0.86$ & $0.96$ & $1.00$ & $1.00$ \\
\bottomrule
\end{tabular}
\caption{Performance of methods across ResNet-9 at $k = 1000$ with $B = 500$ on CIFAR-10, and GPT-2 at $k = 600$ with $B = 300$ on TinyStories. Queries are partitioned into quartiles by $\|Y_q\|$ in increasing order, with Q1 being the smallest and Q4 being the largest. The linear datamodeling score is measured against models retrained after removing a random $1\%$ of the training set. The values are averaged over ten runs for ResNet-9 and four on TinyStories.}
\label{tab:decouple-quartile}
\end{table}

Surprisingly, we also find that the row norms of $Y$ in the scenarios above are nearly perfectly ranked by the norms of the query gradients themselves, with a mean Spearman correlation of $0.999$ on CIFAR-10 and $0.996$ on TinyStories across the same runs. Given that the query gradients are already accessible after training, a practitioner could read off how concentrated a query set is in advance, and decide whether to invoke \textsc{Mage} or \textsc{Spell} before spending any measurement budget. We leave a theoretical justification of why this phenomenon occurs for future investigation. 

\end{document}